\documentclass{article}

\usepackage{times}
\usepackage{graphicx} 
\usepackage{subfigure} 
\usepackage{stmaryrd} 

\usepackage{natbib}

\usepackage{algorithm}
\usepackage{algorithmic}

\usepackage{ftnright} 

\usepackage{hyperref}

\usepackage[accepted]{icml2014}

\usepackage{amsmath}
\usepackage{amssymb}

\icmltitlerunning{Learning the Irreducible Representations of Commutative Lie Groups}

\begin{document} 

\twocolumn[

\icmltitle{Learning the Irreducible Representations of Commutative Lie Groups}

\icmlauthor{Taco Cohen}{T.S.Cohen@uva.nl}
\icmlauthor{Max Welling}{M.Welling@uva.nl}
\icmladdress{Machine Learning Group, University of Amsterdam}

\icmlkeywords{machine learning, representation learning, TSA, Toroidal Subgroup Analysis, Lie groups, invariance}

\vskip 0.3in
]

\begin{abstract}
We present a new probabilistic model of compact commutative Lie groups that produces invariant-equivariant and disentangled representations of data.
To define the notion of disentangling, we borrow a fundamental principle from physics that is used to derive the elementary particles of a system from its symmetries.
Our model employs a newfound Bayesian conjugacy relation that enables 
 fully tractable probabilistic inference over compact commutative Lie groups -- a class that includes the groups that describe the rotation and cyclic translation of images.
We train the model on pairs of transformed image patches, and show that the learned invariant representation is highly effective for classification.
\end{abstract}

\section{Introduction}
\label{sec:introduction}

Recently, the field of deep learning has produced some remarkable breakthroughs.
The hallmark of the deep learning approach is to learn multiple layers of \emph{representation} of data, and much work has gone into the development of representation learning modules such as RBMs and their generalizations \cite{Welling2005}, and autoencoders \cite{Vincent2008}.
However, at this point it is not quite clear what characterizes a good representation.
In this paper, we take a fresh look at the basic principles behind unsupervised representation learning from the perspective of Lie group theory\footnote{
We will at times assume a passing familiarity with Lie groups, but the main ideas of this paper should be accessible to a broad audience.}.

Various desiderata for learned representations have been expressed:
representations should be meaningful \cite{BengioLecunn_ICLR_2014}, invariant \cite{Goodfellow2009}, abstract and disentangled \cite{Bengio2013},
but so far most of these notions have not been defined in a mathematically precise way.
Here we focus on the notions of invariance and disentangling, leaving the search for meaning for future work.

What do we mean, intuitively, when we speak of invariance and disentangling?
A disentangled representation is one that explicitly represents the distinct factors of variation in the data.
For example, visual data (i.e. pixels) can be thought of as a composition of object identity, position and pose, lighting conditions, etc.
Once disentangling is achieved, invariance follows easily:
to build a representation that is invariant to the transformation of a factor of variation (e.g. object position) that is considered a nuisance for a particular task (e.g. object classification), one can simply ignore the units in the representation that encode the nuisance factor.

To get a mathematical handle on the concept of disentangling, we borrow a fundamental principle from physics, which we refer to as Weyl's principle, following Kanatani (\citeyear{Kanatani1990}).
In physics, this idea is used to tease apart (i.e. disentangle) the  \emph{elementary particles} of a physical system from mere measurement values that have no inherent physical significance.
We apply this principle to the area of vision, for after all, pixels are nothing but physical measurements.

Weyl's principle presupposes a symmetry group that acts on the data.
By this we mean a set of transformations that does not change the ``essence'' of the measured phenomenon, although it may change the ``superficial appearance'', i.e. the measurement values.
As a concrete example that we will use throughout this paper, consider the group known as $\textup{SO}(2)$, acting on images by 2D rotation about the origin.
A transformation from this group (a rotation) may change the value of every pixel in the image, but leaves invariant the identity of the imaged object.
Weyl's principle states that the elementary components of this system are given by the \emph{irreducible representations} of the symmetry group -- a concept that will be explained in this paper.

Although this theoretical principle is widely applicable,
we demonstrate it for real-valued compact commutative groups only.
We introduce a probabilistic model that describes a representation of such a group, and show how it can be learned from pairs of images related by arbitrary and unobserved transformations in the group.
Compact commutative groups are also known as toroidal groups,
so we refer to this model as Toroidal Subgroup Analysis (TSA).
Using a novel conjugate prior, the model integrates probability theory and Lie group theory in a very elegant way.
All the relevant probabilistic quantities such as normalization constants, moments, KL-divergences, the posterior density over the transformation group, the marginal density in data space, and their gradients can be obtained in closed form.

\subsection{Related work}

The first to propose a model and algorithm for learning Lie group representations from data were Rao \& Ruderman (\citeyear{Rao1999}).
This model deals only with one-parameter groups, a limitation that was later lifted by Miao and Rao (\citeyear{Miao2007}).
Both works rely on MAP-inference procedures that can only deal with infinitesimally small transformations.
This problem was solved by \mbox{Sohl-Dickstein} et al. (\citeyear{Sohl-Dickstein2010}) using an elegant adaptive smoothing technique, making it possible to learn from large transformations.
This model uses a general linear transformation to diagonalize a one-parameter group, and combines multiple one-parameter groups multiplicatively.

Other, non-group-theoretical approaches to learning transformations and invariant representations exist~\cite{Memisevic2010}.
These \emph{gating models} were found to perform a kind of joint eigenspace analysis~\cite{Memisevic2012}, which is somewhat similar to the irreducible reduction of a toroidal group.

Motivated by a number of statistical phenomena observed in natural images, Cadieu \& Olshausen (\citeyear{Cadieu2012}) describe a model that decomposes a signal into invariant amplitudes and covariant phase variables.

None of the mentioned methods take into account the full uncertainty over transformation parameters, as does TSA.
Due to exact or approximate symmetries in the data, there is in general no unique transformation relating two images, so that only a multimodal posterior distribution over the group gives a complete description of the geometric situation.
Furthermore, posterior inference in our model is performed by a very fast feed-forward procedure, whereas the MAP inference algorithm by Sohl-Dicksteint et al. requires a more expensive iterative optimization.

\section{Preliminaries}

\subsection{Equivalence, Invariance and Reducibility}
\label{sec:equivalence_invariance_and_reducibility}

In this section, we discuss three fundamental concepts on which the analysis in the rest of this paper is based: equivalence, invariance and reducibility.

Consider a function $\Phi : \mathbb{R}^D \rightarrow X$ that assigns to each possible data point $\mathbf{x} \in \mathbb{R}^D$ a class-label ($X = \{1, \ldots, L\}$)
or some distributed representation (e.g. $X = \mathbb{R}^L$).
Such a function induces an equivalence relation on the input space $\mathbb{R}^D$: we say that two vectors $\mathbf{x}, \mathbf{y} \in \mathbb{R}^D$ are $\Phi$-equivalent if they are mapped onto the same representation by $\Phi$.
Symbolically, $\mathbf{x} \equiv_{\Phi} \mathbf{y} \Leftrightarrow \Phi(\mathbf{x}) = \Phi(\mathbf{y})$.

Every equivalence relation on the input space fully determines a \emph{symmetry group} acting on the space.
This group, call it $G$, contains all invertible transformations $\rho : \mathbb{R}^D \rightarrow \mathbb{R}^D$ that leave $\Phi$ invariant: $G = \{ \rho \, |\, \forall \mathbf{x} \in \mathbb{R}^ D : \Phi( \rho(\mathbf{x})) = \Phi(\mathbf{x}) \}$.
$G$ describes the symmetries of $\Phi$, or, stated differently, the label function/representation $\Phi$ is \emph{invariant} to transformations in $G$.
Hence, we can speak of \emph{G-equivalence}: $\mathbf{x} \equiv_G \mathbf{y} \Leftrightarrow \exists \rho \in G : \rho(\mathbf{x}) = \mathbf{y}$.
For example, if some elements of $G$ act by rotating the image, two images are $G$-equivalent if they are rotations of each other.

Before we can introduce Weyl's principle, we need one more concept: the \emph{reduction} of a group representation \cite{Kanatani1990}.
Let us restrict our attention to linear representations of Lie groups: $\rho$ becomes a matrix-valued function $\rho_g$ of an abstract group element $g \in G$, such that $\forall g,h \in G: \rho_{g \circ h} = \rho_g \rho_h$.
In general, every coordinate $y_i$ of $\mathbf{y} = \rho_g \mathbf{x}$ can depend on every coordinate $x_j$ of $\mathbf{x}$. 
Now, since $\mathbf{x}$ is G-equivalent to $\mathbf{y}$,
\emph{it makes no sense to consider the coordinates $x_i$ as separate quantities}; we can only consider the vector $\mathbf{x}$ as a single unit because the symmetry transformations $\rho_g$ tangle all coordinates.
In other words, we cannot say that coordinate $x_i$ is an independent part of the aggregate $\mathbf{x}$, because a mapping $\mathbf{x} \rightarrow \mathbf{x}' = \rho_g \mathbf{x}$ that is supposed to leave the intrinsic properties of $\mathbf{x}$ unchanged, will in fact induce induce a functional dependence between all supposed parts $x_i'$ and $x_j$.

However, we are free to change the basis of the measurement space.
It may be possible to use a change of basis to expose an \emph{invariant subspace}, i.e. a subspace $V \subset \mathbb{R}^D$ that is mapped onto itself by every transformation in the group: $\forall g \in G: \mathbf{x} \in V \Rightarrow \rho_g \mathbf{x} \in V$.
If such a subspace exists and its orthogonal complement $V^{\bot} \subset \mathbb{R}^D$ is also an invariant subspace, then it makes sense to consider the two parts of $\mathbf{x}$ that lie in $V$ and $V^{\bot}$ to be distinct, because they remain distinct under symmetry transformations.

Let $\mathbf{W}$ be a change of basis matrix that exposes the invariant subspaces, that is,
\begin{equation}
  \rho_g = 
    \mathbf{W} 
    \begin{bmatrix}
      \rho_g^1 & \\
       &   \rho_g^2 \\ 
    \end{bmatrix}
    \mathbf{W}^{-1},
\end{equation}
for \emph{all} $g \in G$.
Both $\rho^1_g$ and $\rho_g^2$ form a representation of the \emph{same} abstract group as represented by $\rho$.
The group representations $\rho_g^1$ and $\rho_g^2$ describe how the individual parts $\mathbf{x}_1 \in V$ and $\mathbf{x}_2 \in V^{\bot}$ are transformed by the elements of the group.
As is common in group representation theory, we refer to both the group representations 
$\rho_g^1, \rho_g^2$
and the subspaces $V$ and $V^{\bot}$ corresponding to these group representations as ``representations''.

The process of reduction can be applied recursively to $\rho_g^1$ and $\rho_g^2$.
If at some point there is no more (non-trivial) invariant subspace, the representation is called \emph{irreducible}.
Weyl's principle states that \emph{the elementary components of a system are the irreducible representations of the symmetry group of the system.}
Properly understood, it is not a physics principle at all, but generally applicable to any situation where there is a well-defined notion of equivalence\footnote{The picture becomes a lot more complicated, though, when the group does not act linearly or is not completely reducible}.
It is completely abstract and therefore agnostic about the type of data (images, optical flows, sound, etc.), making it eminently useful for representation learning.

In the rest of this paper, we will demonstrate Weyl's principle in the simple case of a compact commutative subgroup of the special orthogonal group in $D$ dimensions. 
We want to stress though, that there is no reason the basic ideas cannot be applied to non-commutative groups acting on non-linear latent representation spaces.

\subsection{Maximal Tori in the Orthogonal Group}
\label{sec:maximaltori}

In order to facilitate analysis, we will from here on consider only compact commutative subgroups of the special orthogonal group $SO(D)$.
For reasons that will become clear shortly, such groups are called toroidal subgroups of $SO(D)$.
Intuitively, the toroidal subgroups of general compact Lie groups can be thought of as the ``commutative part'' of these groups.
This fact, combined with their analytic tractability (evidenced by the results in this paper) makes them suitable as the starting point of a theory of probabilistic Lie-group representation learning.

Imposing the constraint of orthogonality will make the computation of matrix inverses very cheap, because for orthogonal $\mathbf{Q}$, $\mathbf{Q}^{-1} = \mathbf{Q}^T$.
Orthogonal matrices also avoid numerical problems, because their condition number is always equal to 1.
Another important property of orthogonal transformations is that they leave the Euclidean metric invariant: $\|\mathbf{Qx}\| = \|\mathbf{x}\|$.
Therefore, orthogonal matrices cannot express transformations such as contrast scaling, but they can still model the interesting structural changes in images \cite{Bethge2007}.
For example, since 2D image rotation and (cyclic) translation are linear and do not change the total energy (norm) of the image, they can be represented by orthogonal matrices acting on vectorized images.

As is well known, commuting matrices can be simultaneously diagonalized, so one could represent a toroidal group in terms of a basis of eigenvectors shared by every element in the group, and one diagonal matrix of eigenvalues for each element of the group, as was done in \cite{Sohl-Dickstein2010} for 1-parameter Lie groups.
However, orthogonal matrices do not generally have a complete set of real eigenvectors.
One could use a complex basis instead, but this introduces redundancies because the eigenvalues and eigenvectors of an orthogonal matrix come in complex conjugate pairs.
For machine learning applications, this is clearly an undesirable feature,
so we opt for a joint block-diagonalization of the elements of the toroidal group:
$\rho_\varphi = \mathbf{W} \mathbf{R}(\varphi) \mathbf{W}^T$,
where $\mathbf{W}$ is orthogonal and $\mathbf{R}(\varphi)$ is a block-diagonal rotation matrix\footnote{For ease of exposition, we assume an even dimensional space $D=2J$, but the equations are easily generalized.}:
\begin{equation}
\label{eq:R_block_diag}
\mathbf{R}(\varphi) = 
\begin{bmatrix}
\mathbf{R}(\varphi_1) &  \\
& \ddots &  \\
& & \mathbf{R}(\varphi_J) \\
 \end{bmatrix}.
\end{equation}
The diagonal of $\mathbf{R}(\varphi)$ contains $2\times 2$ rotation matrices 
\begin{equation}
\label{eq:R_block}
\mathbf{R}(\varphi_j) =
\begin{bmatrix}
\cos(\varphi_j) & -\sin(\varphi_j) \\
\sin(\varphi_j) & \cos(\varphi_j)
\end{bmatrix}.
\end{equation}
In this parameterization, the real, orthogonal basis $\mathbf{W}$ identifies the group representation, while the vector of rotation angles $\varphi$ identifies a particular element of the group.
It is now clear why such groups are called ``toroidal'': the parameter space $\varphi$ is periodic in each element $\varphi_j$ and hence is a topological torus.
For a $J$-parameter toroidal group, all the $\varphi_j$ can be chosen freely.
Such a group is known as a \emph{maximal} torus in $SO(D)$, for which we write $\mathbb{T}^J = \{ \varphi \, | \, \varphi_j \in [0, 2 \pi], j=1,\ldots J\}$.

To gain insight into the structure of toroidal groups with fewer parameters, we rewrite eq. \ref{eq:R_block_diag} using the matrix exponential:
\begin{equation}
\mathbf{R}(\varphi) = \exp{\left(\sum_{j=1}^J \varphi_j \mathbf{A}_j \right)}.
\end{equation}
The anti-symmetric matrices $\mathbf{A}_j = \frac{d}{d\varphi_j} \mathbf{R}(\varphi)\big|_{\mathbf{0}}$ are known as the Lie algebra generators, and the $\varphi_j$ are Lie-algebra coordinates.

The Lie algebra is a structure that largely determines the structure of the corresponding Lie group, while having the important advantage of forming a linear space.
That is, all linear combinations of the generators belong to the Lie algebra, and each element of the Lie algebra corresponds to an element of the Lie group, which itself is a non-linear manifold.
Furthermore, every sub\emph{group} of the Lie group corresponds to a sub\emph{algebra} (not defined here) of the Lie algebra.
All toroidal groups are the subgroup of some maximal torus, so we can learn a general toroidal group by first learning a maximal torus and then learning a subalgebra of its Lie algebra.
Due to commutativity, the structure of the Lie algebra of toroidal groups is such that any sub\emph{space} of the Lie algebra is in fact a subalgebra.
The relevance of this observation to our machine learning problem is that \emph{to learn a toroidal group with $I$ parameters ($I < J$), we can simply learn a maximal toroidal group and then learn an $I$-dimensional linear subspace in the space of $\varphi$.}

\begin{figure}
\centering
\includegraphics[scale=0.4]{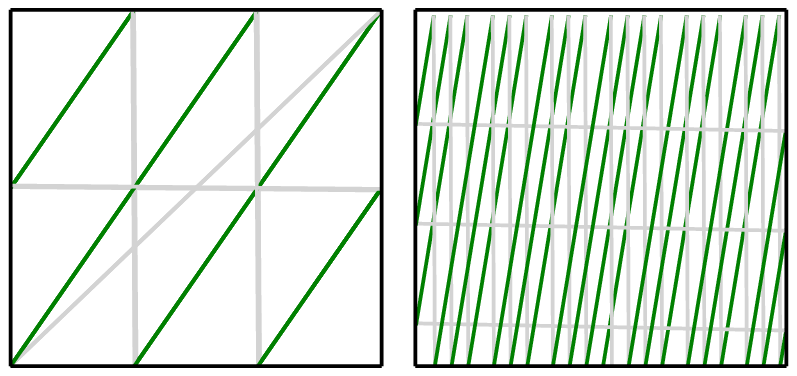}
\caption{Parameter space $\varphi = (\omega_1 s, \omega_2 s)$ of two toroidal subgroups, for $\omega_1 = 2, \omega_2=3$ (left) and $\omega_1=1, \omega_2 = 2\pi$ (right). The point $\varphi$ moves over the dark green line as $s$ is changed. Wrapping around is indicated in light gray. In the incommensurable case (right), coupling does not add structure to the model, because all transformations (points in the plane) can still be constructed by an appropriate choice of $s$.}
\label{fig:toroidal_orbits}
\end{figure}

In this work, we are interested in compact subgroups only\footnote{The main reason for this restriction is that compact groups are simpler and better understood than non-compact groups. In practice, many non-compact groups can be compactified, so not much is lost.}, which is to say that the parameter space should be closed and bounded.
To see that not all subgroups of a maximal torus are compact, consider a 4D space and a maximal torus with 2 generators $\mathbf{A}_1$ and $\mathbf{A}_2$.
Let us define a subalgebra with one generator $\mathbf{A} = \omega_1 \mathbf{A}_1 + \omega_2 \mathbf{A}_2$, for real numbers $\omega_1$ and $\omega_2$.
The group elements generated by this algebra through the exponential map takes the form 
\begin{equation}
\label{eq:R_block_diag_omega}
\mathbf{R}(s) = \exp{(s \mathbf{A})} =
\begin{bmatrix}
\mathbf{R}(\omega_1 s) &  \\
& & \mathbf{R}(\omega_2 s) \\
 \end{bmatrix}.
\end{equation}
Each block $\mathbf{R}(\omega_j s)$ is periodic with period $2 \pi / \omega_j$, but their direct sum $\mathbf{R}(s)$ need not be.
When $\omega_1$ and $\omega_2$ are not commensurate, all values of $s \in \mathbb{R}$ will produce different $\mathbf{R}(s)$, and hence the parameter space is not bounded.
To obtain a compact one-parameter group with parameter space $s \in [0, 2\pi]$, we restrict the frequencies $\omega_j$ to be integers, so that $\mathbf{R}(s) = \mathbf{R}(s + 2\pi)$ (see figure \ref{fig:toroidal_orbits}).

It is easy to see that each block of $\mathbf{R}(s)$ forms a real-valued irreducible representation of the toroidal group, making $\mathbf{R}(s)$ a direct sum of irreducible representations.
From the point of view expounded in section \ref{sec:equivalence_invariance_and_reducibility}, we should view the vector $\mathbf{x}$ as a tangle of \emph{elementary components} $\mathbf{u}_j = \mathbf{W}_j^T \mathbf{x}$, where $\mathbf{W}_j = (\mathbf{W}_{(:, \; 2j - 1)}, \mathbf{W}_{(:, \; 2j)})$ denotes the $D \times 2$ submatrix of $\mathbf{W}$ corresponding to the $j$-th block in $\mathbf{R}(s)$.
Each one of the elementary parts $\mathbf{u}_j$ is functionally independent of the others under symmetry transformations.

The variable $\omega_j$ is known as the \emph{weight} of the representation \cite{Kanatani1990}.
When the representations are equivalent (i.e. they have the same weight), the parts are ``of the same kind'' and are transformed identically.
Elementary components with different weights transform differently.

In the following section, we show how a maximal toroidal group and a 1-parameter subgroup can be learned from correspondence pairs, and how these can be used to generate invariant representations.

\section{Toroidal Subgroup Analysis}

We will start by modelling a maximal torus.
A data pair $(\mathbf{x}, \mathbf{y})$ is related by a transformation
 $\rho_\varphi = \mathbf{W} \mathbf{R}(\varphi) \mathbf{W}^T$:
\begin{equation}
\mathbf{y} = \mathbf{W}\mathbf{R}(\varphi) \mathbf{W}^T \mathbf{x} + \mathbf{\epsilon},
\end{equation}
where $\epsilon \sim \mathcal{N}(\mathbf{0}, \sigma^2)$ represents isotropic Gaussian noise.
In other symbols, $p(\mathbf{y}| \mathbf{x}, \varphi) = \mathcal{N}(\mathbf{y} | \mathbf{W}\mathbf{R}(\varphi)\mathbf{W}^T\mathbf{x}, \sigma^2)$.

We use the following notation for indexing invariant subspaces.
As before, $\mathbf{W}_j = (\mathbf{W}_{(:, \; 2j - 1)}, \mathbf{W}_{(:, \; 2j)})$.
Let $\mathbf{u}_j = \mathbf{W}_j^T \mathbf{x}$ and $\mathbf{v}_j = \mathbf{W}_j^T \mathbf{y}$.
If we want to access one of the coordinates of $\mathbf{u}$ or $\mathbf{v}$, we write $u_{j_1} = 
\mathbf{W}_{(:, \; 2j - 1)}^T \mathbf{x}$ or $u_{j_2} = \mathbf{W}_{(:, \; 2j)}^T \mathbf{x}$.

We assume the $\varphi_j$ to be marginally independent and von-Mises distributed.
The von-Mises distribution is an exponential family that assigns equal density to the endpoints of any length-$2 \pi$ interval of the real line, making it a suitable choice for periodic variables such as $\varphi_j$.
We will find it convenient to move back and forth between the conventional and natural parameterizations of this distribution.
The conventional parameterization of the von-Mises distribution $\mathcal{M}(\varphi_j | \mu_j, \kappa_j)$ uses a mean $\mu_j$ and precision $\kappa_j$:
\begin{equation}
p(\varphi_j) = \frac{1}{2 \pi I_0(\kappa_j)} \exp{(\kappa_j \cos(\varphi_j - \mu_j))}.§
\end{equation}
The function $I_0$ that appears in the normalizing constant is known as the modified Bessel function of order 0.

Since the von-Mises distribution is an exponential family, we can write it in terms of natural parameters $\eta_j = (\eta_{j_1}, \eta_{j_2})^T$ as follows:
\begin{equation}
p(\varphi_j) = \frac{1}{2 \pi I_0(\|\eta_j\|)} \exp{(\eta_j^T T(\varphi_j))},
\end{equation}
where $T(\varphi_j) = (\cos(\varphi_j), \sin(\varphi_j))^T$ are the sufficient statistics.
The natural parameters can be computed from conventional parameters using,
\begin{equation}
\label{eq:conventional_to_natural}
\eta_j = \kappa_j [\cos(\mu_j), \, \sin(\mu_j)]^T
\end{equation}
and vice versa,
\begin{equation}
\label{eq:natural_to_conventional}
\begin{aligned}
\kappa_j &= \|\eta_j\|, &&& \mu_j &= \tan^{-1}(\eta_{j_2} / \eta_{j_1})
\end{aligned}
\end{equation}

Using the natural parameterization, it is easy to see that the prior is conjugate to the likelihood, so that the posterior $p(\varphi | \mathbf{x}, \mathbf{y})$ is again a product of von-Mises distributions.
Such conjugacy relations are of great utility in Bayesian statistics, because they simplify sequential inference.
To our knowledge, this conjugacy relation has not been described before.
To derive this result, first observe that the likelihood term splits into a sum over the invariant subspaces indexed by $j$:
\begin{align*}
p(\varphi | \mathbf{x}, \mathbf{y}) \propto& \, p(\mathbf{y} | \mathbf{x}, \varphi) p(\varphi) \\
\propto& \exp{\left( -\frac{1}{2 \sigma^2} \|\mathbf{y} - \mathbf{W} \mathbf{R}(\varphi) \mathbf{W}^T \mathbf{x}\|^2 \right)} p(\varphi) \\
\propto& \exp{\left(\sum_{j=1}^J \frac{\mathbf{v}_j^T \mathbf{R}(\varphi_j) \mathbf{u}_j}{\sigma^2}  + \eta_j^T T(\varphi_j) \right)} \\
\end{align*}
Both the bilinear forms $\mathbf{v}_j^T \mathbf{R}(\varphi_j) \mathbf{u}_j$ and the prior terms $\eta_j^T T(\varphi)$ are linear functions of $\cos(\varphi_j)$ and $\sin(\varphi_j)$, so that they can be combined into a single dot product:
\begin{equation}
p(\varphi | \mathbf{x}, \mathbf{y}) \propto \exp{\left[ \sum_{j=1}^J \hat{\eta}_j^T T(\varphi_j) \right]},
\end{equation}
which we recognize as a product of von-Mises in natural form.

The parameters $\hat{\eta}_j$ of the posterior are given by:
\begin{equation}
\label{eq:posterior_params}
\begin{aligned}
\hat{\eta}_j &= \eta_j + \frac{1}{\sigma^2} [u_{j_1} v_{j_1} + u_{j_2} v_{j_2}, \, u_{j_1} v_{j_2} - u_{j_2} v_{j_1}]^T \\
&= \eta_j + \frac{\|\mathbf{u}_j\| \|\mathbf{v}_j\|}{\sigma^2} [\cos(\theta_j), \, \sin(\theta_j)]^T,
\end{aligned}
\end{equation}
where $\theta_j$ is the angle between $\mathbf{u}_j$ and $\mathbf{v}_j$.
Geometrically, we can interpret the Bayesian updating procedure in eq. \ref{eq:posterior_params} as follows.
The orientation of the natural parameter vector $\eta_j$ determines the mean of the von-Mises, while its magnitude determines the precision.
To update this parameter with new information obtained from data $\mathbf{u}_j$, $\mathbf{v}_j$, one should add the vector $(\cos(\theta_j), \sin(\theta_j))^T$ to the prior, using a scaling factor that grows with the magnitude of $\mathbf{u}_j$ and $\mathbf{v}_j$ and declines with the square of the noise level $\sigma$.
The longer $\mathbf{u}_j$ and $\mathbf{v}_j$ and the smaller the noise level, the greater the precision of the observation.
This geometrically sensible result follows directly from the consistent application of the rules of probability.

Observe that when using a uniform prior (i.e. $\eta_j=\mathbf{0}$), the posterior mean $\hat{\mu}_j$ (computed from $\hat{\eta}_j$ by eq. \ref{eq:natural_to_conventional}) will be exactly equal to the angle $\theta_j$ between $\mathbf{u}_j$ and $\mathbf{v}_j$.
We will use this fact in section \ref{sec:invariant_representation_and_metric} when we derive the formula for the orbit distance in a toroidal group.

Previous approaches to Lie group learning only provide point estimates of the transformation parameters, which have to be obtained using an iterative optimization procedure \cite{Sohl-Dickstein2010}.
In contrast, TSA provides a full posterior distribution which is obtained using a simple feed-forward computation.
Compared to the work of Cadieu \& Olshausen (\citeyear{Cadieu2012}), our model deals well with low-energy subspaces, by simply describing the uncertainty in the estimate instead of providing inaccurate estimates that have to be discarded.

\subsection{Invariant Representation and Metric}
\label{sec:invariant_representation_and_metric}

One way of doing invariant classification is by using an invariant metric known as the \emph{manifold distance}.
This metric $d(\mathbf{x}, \mathbf{y})$ is defined as the minimum distance between the orbits
$O_\mathbf{x} = \{\rho_\varphi \mathbf{x}\ | \varphi \in G\}$ and $O_\mathbf{y} = \{\rho_\varphi \mathbf{y}\ | \varphi \in G\}$.
Observe that this is only a true metric that satisfies the coincidence axiom $d(\mathbf{x}, \mathbf{y}) = 0 \Leftrightarrow \mathbf{x} = \mathbf{y}$ if we take the condition $\mathbf{x}=\mathbf{y}$ to mean ``equivalence up to symmetry transformations'' or $\mathbf{x} \equiv_G \mathbf{y}$, as discussed in section \ref{sec:equivalence_invariance_and_reducibility}.

In practice, it has proven difficult to compute this distance exactly, so approximations such as tangent distance have been invented~\cite{Simard2000}.
But for a maximal torus, we can easily compute the exact manifold distance:
\begin{equation}
\label{eq:manifold_dist_maximal_torus}
\begin{aligned}
d^2(\mathbf{x}, \mathbf{y}) &= \min_{\varphi} \|\mathbf{y} - \mathbf{W} \mathbf{R}(\varphi)\mathbf{W}^T \mathbf{x} \|^2 \\
&= \sum_j \min_{\varphi_j} \|\mathbf{v}_j - \mathbf{R}(\varphi_j)\mathbf{u}_j\|^2 \\
&= \sum_j \|\mathbf{v}_j - \mathbf{R}(\hat{\mu}_j)\mathbf{u}_j\|^2,
\end{aligned}
\end{equation}
where $\hat{\mu}_j$ is the mean of the posterior $p(\varphi_j | \mathbf{x}, \mathbf{y})$, obtained using a uniform prior ($\kappa_j = 0$).
The last step of eq. \ref{eq:manifold_dist_maximal_torus} follows, because as we saw in the previous section, $\hat{\mu}_j$ is simply the angle between $\mathbf{u}_j$ and $\mathbf{v}_j$ when using a uniform prior.
Therefore, $\mathbf{R}(\hat{\mu}_j)$ aligns $\mathbf{u}_j$ and $\mathbf{v}_j$, minimizing the distance between them.

Another approach to invariant classification is through an invariant representation.
Although the model presented above aims to describe the transformation between observations $\mathbf{x}$ and $\mathbf{y}$, an invariant-equivariant representation appears automatically in terms of the parameters of the posterior over the group.
To see this, consider all the transformations in the learned toroidal group $G$ that take an image $\mathbf{x}$ to itself.
This set is known as the stabilizer $\verb+stab+_G(\mathbf{x})$ of $\mathbf{x}$.
It is a subgroup of $G$ and describes the symmetries of $\mathbf{x}$ with respect to $G$.
When a transformation $\varphi \in G$ is applied to $\mathbf{x}$, the stabilizer subgroup is left invariant, for if 
$\theta \in \verb+stab+_G(\mathbf{x})$ then 
$\rho_\theta \rho_\varphi \mathbf{x} = \rho_\varphi \rho_\theta \mathbf{x} = \rho_\varphi\mathbf{x}$ and hence $\rho_\theta \in \verb+stab+_G(\rho_\varphi\mathbf{x})$.

The posterior of $\mathbf{x}$ transformed into itself, $p(\varphi | \mathbf{x}, \mathbf{x}, \mu, \kappa = \mathbf{0}) = \prod_j \mathcal{M}(\varphi_j | \hat{\mu}_j, \hat{\kappa}_j)$ gives a probabilistic description of the stabilizer of $\mathbf{x}$, and hence must be invariant.
Clearly, the angle between $\mathbf{x}$ and $\mathbf{x}$ is zero, so $\hat{\mu} = \mathbf{0}$.
On the other hand, $\hat{\kappa}$ contains information about $\mathbf{x}$ and is invariant.
To see this, recall that $\hat{\kappa}_j = \|\hat{\eta}_j\|$.
Using eq. \ref{eq:posterior_params} we obtain $\hat{\kappa}_j = \|\mathbf{u}_j\|^2 \sigma^{-2} = \|\mathbf{W}_j^T \mathbf{x}\|^2 \sigma^{-2}$.
Since every transformation $\varphi$ in the toroidal group acts on the 2D vector $\mathbf{u}_j$ by rotation, the norm of $\mathbf{u}_j$ is left invariant.

We recognize the computation of $\hat{\kappa}$ as the \emph{square pooling} operation often applied in convolutional networks to gain invariance: project an image onto filters $\mathbf{W}_{:, 2j -1}$ and $\mathbf{W}_{:, 2j}$ and sum the squares.
This computation is a direct consequence of our model setup.
In section \ref{sec:modeling_a_lie_subalgebra}, we will find that the model for non-maximal tori is even more informative about the proper pooling scheme.

Since we want to use $\hat{\kappa}$ as an invariant representation, we should try to find an appropriate metric on $\hat{\kappa}$-space.
Let $\hat{\kappa}(\mathbf{x})$ be defined by $p(\varphi | \mathbf{x}, \mathbf{x}, \kappa = \mathbf{0}) = \prod_j \mathcal{M}(\varphi | \hat{\mu}_j, \hat{\kappa}_j(\mathbf{x}))$.
We suggest using the Hellinger distance:
\begin{align*}
H^2(\hat{\kappa}(\mathbf{x}), \hat{\kappa}(\mathbf{y}))
&= \frac{1}{2} \sum_j \left(\sqrt{\hat{\kappa}_j(\mathbf{x})} - \sqrt{\hat{\kappa}_j(\mathbf{y})}\right)^2 \\
&= \frac{1}{2 \sigma^2} \sum_j \|\mathbf{u}_j\|^2 + \|\mathbf{v}_j\|^2 - 2\|\mathbf{u}_j\|\|\mathbf{v}_j\| \\
&= \frac{1}{2 \sigma^2} \sum_j \|\mathbf{v}_j - \mathbf{R}(\hat{\mu}_j)\mathbf{u}_j\|^2,
\end{align*}
which is equal to the \emph{exact manifold distance} (eq. \ref{eq:manifold_dist_maximal_torus}) up to a factor of $\frac{1}{2 \sigma^2}$.
The first step of this derivation uses eq. \ref{eq:posterior_params} under a uniform prior ($\eta_j = \mathbf{0}$),
while the second step again makes use of the fact that $\hat{\mu}_j$ is the angle between $\mathbf{u}_j$ and $\mathbf{v}_j$ so that $\|\mathbf{u}_j\|\|\mathbf{v}_j\| = \mathbf{u}_j^T \mathbf{R}(\hat{\mu}_j) \mathbf{v}_j$.

\subsection{Relation to the Discrete Fourier Transform}

We show that the DFT is a special case of TSA.
The DFT of a discrete 1D signal $\mathbf{x} = (x_1, \ldots, x_D)^T$ is defined:
\begin{equation}
X_j = \sum_{n=0}^{D-1} x_n \rho^{-j n}
\end{equation}
where $\rho = e^{2 \pi i / D}$ is the $D$-th primitive root of unity.
If we choose a basis of sinusoids for the filters in $\mathbf{W}$,
\begin{equation*}
\begin{aligned}
\mathbf{W}_{(:, \; 2j-1)} &= \mathfrak{R}(\rho^{-j}, \ldots, \rho^{-j(D-1)})^T \\
 &= (\cos(2 \pi j / D), \ldots, \cos(2 \pi j (D-1)/D))^T \\
\mathbf{W}_{(:, \; 2j)} &= \mathfrak{I}(\rho^{-j}, \ldots, \rho^{-j(D-1)})^T \\
 & = (\sin(-2 \pi j / D), \ldots, \sin(-2 \pi j (D-1)/D))^T,
\end{aligned}
\end{equation*}
  then the change of basis performed by $\mathbf{W}$ is a DFT.
Specifically, $\mathfrak{R}(X_j) = \mathbf{u}_{j_1}$ and $\mathfrak{I}(X_j) = \mathbf{u}_{j_2}$.

Now suppose we are interested in the transformation taking some arbitrary fixed vector $\mathbf{e} = \mathbf{W}(1, 0, \ldots, 1, 0)^T$ to $\mathbf{x}$.
The posterior over $\varphi_j$ is $p(\varphi_j | \mathbf{e}, \mathbf{x}, \eta_j = \mathbf{0}, \sigma=1) = \mathcal{M}(\varphi_j | \hat{\eta}_j)$, where (by eq. \ref{eq:posterior_params}) we have $\hat{\eta}_j = \|\mathbf{u}_j\| [\cos(\theta_j), \sin(\theta_j)]^T$, $\theta_j$ being the angle between $\mathbf{u}_j$ and the ``real axis'' $\mathbf{e}_j = (1, 0)^T$.
In conventional coordinates, the precision of the posterior is equal to the modulus of the DFT, $\hat{\kappa}_j = \|\mathbf{u}_j\| = |X_j|$, and the mean of the posterior is equal to the phase of the Fourier transform, $\hat{\mu} = \theta_j = \arg(X_j)$.
Therefore, TSA provides a probabilistic interpretation of the DFT coefficients, and makes it possible to learn an appropriate generalized transform from data.

\subsection{Modeling a Lie subalgebra}
\label{sec:modeling_a_lie_subalgebra}

Typically, one is interested in learning groups with fewer than $J$ degrees of freedom.
As we have seen, for one parameter compact subgroups of a maximal torus, the weights of the irreducible representations must be integers.
We model this using a coupled rotation matrix, as follows:
\begin{equation}
\label{eq:block_diag_omega_s1}
\rho_s
= \mathbf{W}
\begin{bmatrix}
\mathbf{R}(\omega_1 s) &  \\
& \ddots &  \\
& & \mathbf{R}(\omega_J s)
 \end{bmatrix}
\mathbf{W}^T
\end{equation}
Where $s \in [0, 2\pi]$ is the scalar parameter of this subgroup.
The likelihood then becomes 
$\mathbf{y} \sim \mathcal{N}(\mathbf{y} | \rho_s \mathbf{x}, \sigma^2)$.

We have found that the conjugate prior for this likelihood is the \emph{generalized} von-Mises \cite{Gatto2007}:
\begin{align*}
p(s) = \mathcal{M}^{+}(s | \eta^{+}) &= \exp{\left(\eta^{+} \cdot T^{+}(s) \right)} \; / \; Z^{+}\\
&= \exp{\left(\sum_{j=1}^{K} \kappa_j^{+} \cos(j s - \mu_j^{+})\right)} \; / \; Z^{+}
\end{align*}
where $T^{+}(s) = [\cos(s), \sin(s), \ldots, \cos(K s), \sin(K s)]^T$.

This conjugacy relation $p(s | \mathbf{x}, \mathbf{y}) \propto \exp{(\hat{\eta}^{+} \cdot T^{+}(s))}$ is obtained using similar reasoning as before, yielding the update equation,
\begin{equation}
\begin{aligned}
\hat{\eta}_j^{+} = \eta_j^{+} + \sum_{k: \omega_k = j} \hat{\eta}_k
\end{aligned}
\end{equation}
where $\hat{\eta}_k$ is obtained from eq. \ref{eq:posterior_params} using a uniform prior $\eta_k = \mathbf{0}$.
The sum in this update equation performs a pooling operation over a \emph{weight space}, which is defined as the span of those invariant subspaces $k$ whose weight $\omega_k = j$.
As it turns out, this is exactly the right thing to do in order to summarize the data while maintaining invariance.
As explained by Kanatani (\citeyear{Kanatani1990}), the norm $\|\hat{\eta}^+_j\|$ of a linear combination of same-weight representations is always invariant (as is the maximum of any two $\|\hat{\eta}_j\|$ or $\|\hat{\eta}^{+}_j\|$).
The similarity to sum-pooling and max-pooling in convnets is quite striking.

In the maximal torus model, there are $J=D/2$ degrees of freedom in the group and the invariant representation is $D-J = J$-dimensional ($\hat{\kappa}_1,\ldots,\hat{\kappa}_J$).
This representation of a datum $\mathbf{x}$ identifies a toroidal orbit, and hence the vector $\hat{\kappa}$ is a \emph{maximal invariant} with respect to this group~\cite{Soatto2009}.
For the coupled model, there is only one degree of freedom in the group,
so the invariant representation should be $D - 1$ dimensional.
When all $\omega_k$ are distinct, we have $J$ variables $\kappa_1^{+}, \ldots, \kappa_J^{+}$ that are invariant.
Furthermore, from eq. \ref{eq:block_diag_omega_s1} we see that as $\mathbf{x}$ is transformed,
the angle between $\mathbf{x}$ and an arbitrary fixed reference vector in subspace $j$ transforms as $\theta_j(s) = \delta_j + \omega_j s$ for some data-dependent initial phase $\delta_j$.
It follows that $\omega_j \theta_k(s) - \omega_k \theta_j(s) = \omega_j (\delta_k + \omega_k s) - \omega_k ( \delta_j + \omega_j s ) = \omega_j \delta_k - \omega_k \delta_j$ is invariant.
In this way, we can easily construct another $J-1$ invariants, but unfortunately these are not stable because the angle estimates can be inaccurate for low-energy subspaces.
Finding a stable maximal invariant, and doing so in a way that will generalize to other groups is an interesting problem for future work.

The normalization constant $Z^{+}$ for the GvM has so far only been described for the case of $K=2$ harmonics, but we have found a closed form solution in terms of the so-called modified Generalized Bessel Functions (GBF) of $K$-variables $\kappa^{+} = \kappa_1^{+}, \ldots, \kappa_K^{+}$ and parameters\footnote{As described in the supplementary material, we use a slightly different parameterization of the GBF.} $\exp{(-i \mu^{+})} = \exp{(-i \mu_1^{+})}, \ldots, \exp{(-i \mu_K^{+})}$ \cite{Dattoli1991a}:
\begin{equation}
\label{eq:GvM_Z}
Z^{+}(\kappa^{+}, \mu^{+}) = 2\pi I_0(\kappa^{+}; \; e^{-i \mu^{+}}).
\end{equation}
We have developed a novel, highly scalable algorithm for the computation of GBF of many variables, which is described in the supplementary material.

\begin{figure}
\centering
\includegraphics[scale=0.7]{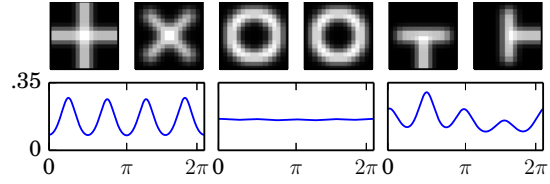}
\caption{Posterior distribution over $s$ for three image pairs.}
\label{fig:posteriors}
\end{figure}
Figure \ref{fig:posteriors} shows the posterior over $s$ for three image pairs related by different rotations and containing different symmetries.
The weights $\mathbf{W}$ and $\omega$ were learned by the procedure described in the next section. 
It is quite clear from this figure that MAP inference does not give a complete description of the possible transformations relating the images when the images have a degree of rotational symmetry.
The posterior distribution of our model provides a sensible way to deal with this kind of uncertainty, which (in the case of 2D translations) is at the heart of the well known aperture problem in vision.
Having a tractable posterior is particularly important if the model is to be used to estimate longer sequences (akin to HMM/LDS models, but non-linear), where one may encounter multiple high-density trajectories.

If required, accurate MAP inference can be performed using the algorithm of Sohl-Dickstein et al. (\citeyear{Sohl-Dickstein2010}), as described in the supplementary material.
This allows us to compute the exact manifold distance for the coupled model.

\subsection{Maximum Marginal Likelihood Learning}

We train the model by gradient descent on the marginal likelihood.
Perhaps surprisingly given the non-linearities in the model, the integrations required for the evaluation of the marginal likelihood can be obtained in closed form for both the coupled and decoupled models.
For the decoupled model we obtain:
\begin{equation}
\label{eq:marginal_likelihood_uncoupled}
\begin{aligned}
p(\mathbf{y} | \mathbf{x}) &= \int_{\varphi \in \mathbb{T}^J} \mathcal{N}(\mathbf{y} | \mathbf{W R}(\varphi) \mathbf{W}^T \mathbf{x}) \prod_j \mathcal{M}(\varphi_j | \eta_j) d\varphi \\
&=  \frac{\exp{\left(-\frac{1}{2\sigma^2} (\|\mathbf{x}\|^2 + \|\mathbf{y}\|^2)\right)}}{\sqrt{(2 \pi \sigma)^D}} \prod_j \frac{I_0(\hat{\kappa}_j)}{I_0(\kappa_j)}.
\end{aligned}
\end{equation}

Observing that $I_0(\hat{\kappa}_j) / I_0(\kappa_j)$ is the ratio of normalization constants of regular von-Mises distributions, the analogous expression for the coupled model is easily seen to be equal to eq. \ref{eq:marginal_likelihood_uncoupled}, only replacing $\prod_j I_0(\hat{\kappa}_j) \; / \; I_0(\kappa_j)$ by $Z^{+}(\hat{\kappa}^{+}, \hat{\mu}^{+}) / Z^{+}(\kappa^{+}, \mu^{+})$.
The derivation of this result can be found in the supplementary material.

The gradient of the log marginal likelihood of the uncoupled model w.r.t. a batch $\mathbf{X}, \mathbf{Y}$ (both storing $N$ vectors in the columns) is:
\begin{align*}
\frac{d}{d\mathbf{W}} \ln p(\mathbf{Y} | \mathbf{X}) &=
\mathbf{X} (\mathbf{R}^T(\mu) \mathbf{W}^T \mathbf{Y} \Yright \mathbf{A}
+  \mathbf{W}^T\mathbf{X} \Yright \mathbf{B}^y)^T \\
&+ \mathbf{Y} (\mathbf{R}(\mu) \mathbf{W}^T \mathbf{X} \Yright \mathbf{A}
+  \mathbf{W}^T\mathbf{Y} \Yright \mathbf{B}^x)^T.
\end{align*}
where we have used $(\mathbf{P} \Yright \mathbf{Q})_{2j, n} = \mathbf{Q}_{j, n} \mathbf{P}_{2j, n}$ and $(\mathbf{P} \Yright \mathbf{Q})_{2j-1, n} = \mathbf{Q}_{j, n} \mathbf{P}_{2j-1, n}$ as a ``subspace weighting'' operation.
$\mathbf{A}$, $\mathbf{B}^{(x)}$ and $\mathbf{B}^{(y)}$ are $D \times N$ matrices with elements
\begin{align*}
a_{j n} &=  \frac{I_1(\hat{\kappa}_{jn}) \kappa_j}
	            {I_0(\hat{\kappa}_{jn}) \hat{\kappa}_{jn} \sigma^2}, \\
b_{j n} &= \frac{\|\mathbf{W}_j \mathbf{y}^{(n)} \|^2}{\kappa_j \sigma^2},
\end{align*}
where the $\hat{\kappa}_{jn}$ is the posterior precision in subspace $j$ for image pair $\mathbf{x}^{(n)}$, $\mathbf{y}^{(n)}$ (the $n$-th column of $\mathbf{X}$, resp. $\mathbf{Y}$).

The gradient of the coupled model is easily computed using the differential recurrence relations that hold for the GBF \cite{Dattoli1991a}.

We use minibatch Stochastic Gradient Descent (SGD) on the log-likelihood of the uncoupled model.
After every parameter update, we orthogonalize $\mathbf{W}$ by setting all singular values to $1$:
Let $\mathbf{U}, \mathbf{S}, \mathbf{V} = \verb+svd+(\mathbf{W})$, then set $\mathbf{W} := \mathbf{U} \mathbf{V}$.
This procedure and all previous derivations still work when the basis is undercomplete, i.e. has fewer columns (filters) than rows (dimensions in data space).
To learn $\omega_j$, we estimate the relative angular velocity $\omega_j = \theta_j / \delta$ from a batch of image patches rotated by a sub-pixel amount $\delta = 0.1^\circ$.

\section{Experiments}
\label{sec:experiments}

\begin{figure}
\begin{center}
\includegraphics[scale=0.5]{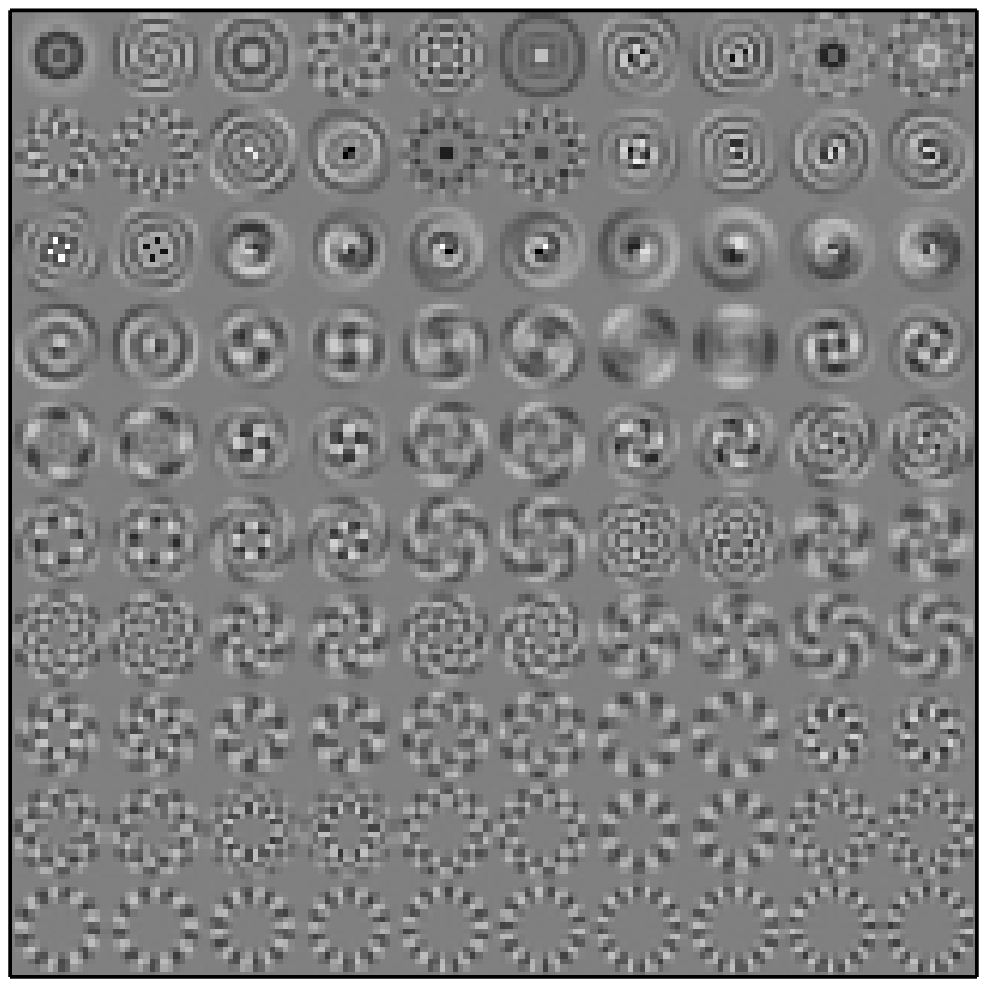}
\end{center}
\caption{Filters learned by TSA, sorted by absolute frequency $|\omega_j|$. The learned $\omega_j$-values range from from $-11$ to $12$.}
\label{fig:filters}
\end{figure}

We trained a TSA model with $100$ filters on a stream of $250.000$ $16\times 16$ image patches $\mathbf{x}^{(t)}$, $\mathbf{y}^{(t)}$.
The patches $\mathbf{x}^{(t)}$ were drawn from a standard normal distribution, and $\mathbf{y}^{(t)}$ was obtained by rotating $\mathbf{x}^{(t)}$ by an angle $s$ drawn uniformly at random from $[0, 2\pi]$.
The learning rate $\alpha$ was initialized at $\alpha_0 = 0.25$ and decayed as $\alpha = \alpha_0 / \sqrt{T}$, where $T$ was incremented by one with each pass through the data.
Each minibatch consisted of $100$ data pairs.
After learning $\mathbf{W}$, we estimate the weights $\omega_j$ and sort the filter pairs by increasing absolute value for visualization.
As can be seen in fig. \ref{fig:filters}, the filters are very clean and the weights are estimated correctly except for a few filters on row $1$ and $2$ that are assigned weight $0$ when in fact they have a higher frequency.

\begin{figure}
\begin{center}
\includegraphics[scale=.7]{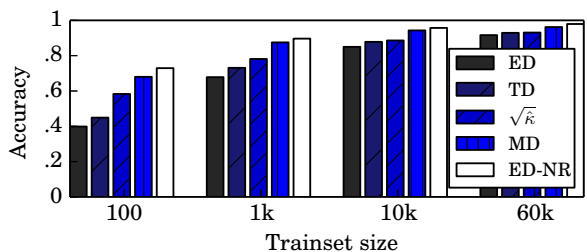}
\end{center}
\caption{Results of classification experiment. See text for details.}
\label{fig:classification_accuracy}
\end{figure}

We tested the utility of the model for invariant classification on a rotated version of the MNIST dataset, using a 1-Nearest Neighbor classifier.
Each digit was rotated by a random angle and rescaled to $16\times 16$ pixels, resulting in 60k training examples and 10k testing examples, with no rotated duplicates.
We compared the Euclidean distance (ED) in pixel space, tangent distance (TD) \cite{Simard2000}, Euclidean distance on the space of $\sqrt{\hat{\kappa}}$ (equivalent to the exact manifold distance for the maximal torus, see section \ref{sec:invariant_representation_and_metric}), the true manifold distance for the 1-parameter 2D rotation group (MD), and the Euclidean distance on the non-rotated version of the MNIST dataset (ED-NR).
The results in fig. \ref{fig:classification_accuracy} show that TD outperforms ED, but is outperformed by $\sqrt{\hat{\kappa}}$ and MD by a large margin.
In fact, the MD-classifier is about as accurate as ED on a much simpler dataset, demonstrating that it has almost completely modded out the variation caused by rotation.

\section{Conclusions and outlook}

We have presented a novel principle for learning disentangled representations, and worked out its consequences for a simple type of symmetry group.
This leads to a completely tractable  model with potential applications to invariant classification and Bayesian estimation of motion.
The model reproduces the pooling operations used in convolutional networks from probabilistic and Lie-group theoretic principles, and provides a probabilistic interpretation of the DFT and its generalizations.

The type of disentangling obtained in this paper is contingent upon the rather minimalist assumption that all that can be said about images is that they are equivalent (rotated copies) or inequivalent.
However, the universal nature of Weyl's principle bodes well for future applications to deep, non-linear and non-commutative forms of disentangling.

\section*{Acknowledgments}
We would like to thank prof. Giuseppe Dattoli for help with derivations related to the generalized Bessel functions.

\bibliography{TSA_ICML}
\bibliographystyle{icml2014}

\end{document}